%% file: main.tex
\documentclass{article}

\usepackage[a4paper, total={6in, 8in}]{geometry}
\usepackage[utf8]{inputenc} 
\usepackage[T1]{fontenc}    
\usepackage{hyperref}       
\usepackage{url}            
\usepackage{booktabs}       
\usepackage{nicefrac}       
\usepackage{microtype}      
\usepackage{xcolor}         
\usepackage{algorithmic}
\usepackage{diagbox}
\usepackage[titlenumbered,ruled]{algorithm2e}
\usepackage{amsmath}
\usepackage{amsthm}
\usepackage{dsfont}
\usepackage{bbm}
\usepackage{subcaption}
\usepackage{graphicx}
\usepackage{multirow}
\usepackage{wrapfig}
\usepackage{comment}
\usepackage{float}
\usepackage{authblk}

\usepackage[style=numeric-comp,useprefix,hyperref,backend=biber,sorting=none]{biblatex} 

\newcommand{\be}{\begin{equation}}
\newcommand{\ee}{\end{equation}}
\newcommand{\bal}{\begin{aligned}}
\newcommand{\eal}{\end{aligned}}

\DeclareMathOperator*{\argmin}{arg\,min}

\newcommand{\scriptR}{\mathcal{R}}
\newcommand{\scriptD}{\mathcal{D}}

\newcommand{\Reals}{\mathds{R}}

\title{\textbf{Learning to Validate Generative Models:\\ a Goodness-of-Fit Approach}}

\author[1,2,3]{Pietro Cappelli\thanks{pietro.cappelli@cern.ch}}
\author[4,5,6]{Gaia Grosso\thanks{gaiag795@mit.edu}}
\author[7,8]{Marco Letizia\thanks{marco.letizia@edu.unige.it}}
\author[9]{Humberto Reyes-González\thanks{humberto.reyes@rwth-aachen.de}}
\author[1,2]{Marco Zanetti\thanks{marco.zanetti@cern.ch}}

\affil[1]{Dipartimento di Fisica e Astronomia, Università di Padova, Padova, Italy}
\affil[2]{INFN, Sezione di Padova, Padova, Italy}
\affil[3]{Department of Physics, Brandeis University, Waltham, MA, USA}
\affil[4]{MIT Laboratory for Nuclear Science, Cambridge, MA, USA}
\affil[5]{School of Engineering and Applied Sciences, Harvard University, Cambridge, MA, USA}
\affil[6]{NSF AI Institute for Artificial Intelligence and Fundamental Interactions, Cambridge, MA, USA}
\affil[7]{MaLGa–DIBRIS, Università di Genova, Genova, Italy}
\affil[8]{INFN, Sezione di Genova, Genova, Italy}
\affil[9]{Institute for Theoretical Particle Physics and Cosmology, RWTH Aachen University, Aachen, Germany}

\begin{document}

\maketitle

\begin{abstract}
\normalsize
Generative models are increasingly central to scientific workflows, yet their systematic use and interpretation require a proper understanding of their limitations through rigorous validation. Classic approaches struggle with scalability, statistical power, or interpretability when applied to high-dimensional data, making it difficult to certify the reliability of these models in realistic, high-dimensional scientific settings. Here, we propose the use of the New Physics Learning Machine (NPLM), a learning-based approach to goodness-of-fit testing inspired by the Neyman--Pearson construction, to test generative networks trained on high-dimensional scientific data. We demonstrate the performance of NPLM for validation in two benchmark cases: generative models trained on mixtures of Gaussian models with increasing dimensionality, and a public end-to-end model, known as \emph{FlowSim}, developed to generate high-energy physics collision events. We demonstrate that the NPLM can serve as a powerful validation method while also providing a means to diagnose sub-optimally modeled regions of the data.

\end{abstract}
\date{}
\newpage
\tableofcontents

\section{Introduction}\label{sec:intro}
\input{Introduction}

\section{Evaluating generative models via goodness-of-fit testing} \label{sec:NPLM}
\input{Methodology}

\section{Experiments}\label{sec:Experiments}
\input{MoG_NFs}\label{subsec:MoG}
\input{FlashSim}\label{subsec:flashsim}

\section{Diagnostics}\label{sec:Diagnosis}
\input{Diagnosis}

\section{Conclusions}\label{sec:Conclusions}

In this work, we investigate the effectiveness of the NPLM goodness-of-fit test for evaluating and validating generative networks trained on complex scientific data. We demonstrate its performance in two benchmark settings: a synthetic case based on mixtures of Gaussian distributions with increasing dimensionality, and a real-world case involving LHC jet data. Our study shows that this methodology provides a statistically robust framework for this task and that the test output correlates strongly with the quality of the generative network. In general, we find that the framework is powerful enough to expose mismodelling even in state-of-the-art generative networks, using relatively small test data samples.

We show that the NPLM test can be used to set an upper bound on the number of events that can be generated before potential mismodelings of the generative models become apparent. A key advantage here is that the NPLM test is inherently multivariate and anomaly-agnostic. While it is generally convenient to take the true data distribution as the reference for the test, thus defining a universal null hypothesis against which any generative model can be evaluated, we also investigated the complementary case where the generative model itself is used as the reference, assessing how well it fits a set of true data. This approach may be useful when true test data are too scarce, although the whole test must be repeated for every generative model. We find that using generated data as the reference leads to less stringent results. A possible explanation is that generative models typically produce samples with support spanning a broader region of the parameter space than the true data distribution. Generally, tests on imbalanced datasets tend to yield asymmetric outcomes, as the information available to the test is inherently uneven. 

Furthermore, we show that the NPLM method can also serve as a diagnostic tool for characterizing anomalous features in generative models. This is achieved by leveraging the underlying classifier of the algorithm to assign anomaly scores to individual data points. By examining the distribution of these predictions, mismodelings in the generative models can be identified. In addition, analyzing discrepant events in marginal or two-dimensional distributions highlights regions where the network underperforms, offering practical guidance for improving generative architectures. Future research will investigate how to maximally exploit this feature, for instance through low-dimensional latent representations.

As an outlook, a ready-to-use implementation of our method is currently in preparation for public release. We are also developing distributions under the null hypothesis for benchmark datasets in HEP. The goal is to provide a robust, quantitative framework for assessing the accuracy of generative networks in reproducing HEP data, as well as for determining the level of statistical confidence with which they can be reliably used as surrogates. Finally, we stress that the proposed approach is general and can be applied to generative models in any domain where the model defines a probability density, either explicitly or implicitly.
\section*{Code availability}
The code used for the MoG study is provided in Ref.~\cite{NPLM-Flows}, while the trained NF models and MoG implementations are available in Ref.~\cite{NF_MoGs_NPLM}. The code employed in the FlowSim study can be found in Ref.~\cite{NPLM-FlashSim}.

\section*{Acknowledgements}
The work of P.C. was supported by ICSC – Centro Nazionale di Ricerca in High Performance Computing, Big Data and Quantum Computing, funded by European Union – NextGenerationEU. G.G. acknowledges the financial support of the National Science Foundation under Cooperative Agreement PHY-2019786 (The NSF AI Institute for Artificial Intelligence and Fundamental Interactions, http://iaifi.org/). M.L. acknowledges the financial support of the European Research Council (grant SLING 819789). The research of H.R.G. is supported by the German Research Foundation DFG under grant 396021762 -- TRR 257: Particle physics phenomenology after the Higgs discovery. H.R.G. and G.G. thank the organizers of the 2025 MIAPbP workshop “Build Big or Build Smart: Examining Scale and Domain Knowledge in Machine Learning for Fundamental Physics”. M.L. thanks the Institute for Theoretical Particle Physics and Cosmology at RWTH Aachen University, through DFG grant 396021762, for providing the opportunity to discuss this project.

\printbibliography

\newpage

\appendix
\section{NPLM model selection}\label{app:modelselection}
\input{nplm}

\end{document}

%% file: Introduction.tex
Generative artificial intelligence (AI) tools have drawn significant attention in scientific research, where they have demonstrated the potential to reshape the discovery workflow and have already contributed to advances in fields such as protein design, drug discovery, and materials science (see for example \cite{NEURIPS2019_f3a4ff48,anstine2023generative,zeni2025generative}). In the path to scientific discovery, the modeling of probability densities plays a crucial role to address very complex theory calculations, parameter estimation and perform data analysis. Recent developments have introduced new highly expressive generative models and sophisticated training strategies able to capture the non-trivial structures characterizing scientific data, providing surrogate models that can be sampled efficiently. In the field of physical sciences, generative AI is providing the tools to accelerate theory calculations and numerical simulations of physical phenomena, opening the way to the statistical interpretations of very complex systems otherwise hard to handle. For instance, in high-energy physics (HEP), fast simulations could help overcome the limitations of computational and storage resources anticipated in the future high-luminosity phase of the Large Hadron Collider (LHC) \cite{ZurbanoFernandez:2020cco}. Therefore generative networks (GNs) are being applied to a variety of complex tasks, such as LHC event generation \cite{Butter:2022rso}, calorimeter shower simulation \cite{Krause_2025}, anomaly detection \cite{Kasieczka:2021xcg}, likelihood modeling \cite{Reyes-Gonzalez:2023oei,Heimel:2024drk}, and more.

While these developments are very promising, the degree to which generative techniques can replace traditional high-fidelity methods for data simulation and analyses depends on the level of precision and control required for specific applications. Alongside improving the sophistication of generative models, it is therefore crucial to develop competitive tools to test their goodness and quantify the extent to which they can be trusted. For example, in HEP, synthetic data are employed to estimate the expected likelihood of the observed data under different theories and experimental setups. This likelihood is then used to estimate theory parameters and experimental conditions or to make scientific discoveries. In such cases, the accuracy of the faithfulness of the data produced by a generative model is of foremost importance for making reliable statements.

The validation of generative models is often performed using some notion of distance between probability measures, such as the Wasserstein distance \cite{villani2008optimal} or the Kolmogorov--Smirnov distance. The use of metrics computed on the output of machine learning (ML) classifiers has become a popular trend in recent years due to their expressiveness and ability to capture intricate correlation patterns in high-dimensional settings (examples in HEP include Refs. \cite{Das:2023ktd,Kansal:2022spb,Ahmad:2024dql}). However, in order to obtain statistically robust responses from these techniques, they need to be deployed in the framework of statistical hypothesis testing. This permits the computation of p-values that can be compared to pre-determined significance levels to quantitatively asses how well surrogate models reproduce the true data distribution, while accounting for statistical fluctuations. 

In this work, we adopt the New Physics Learning Machine (NPLM), a machine learning-based tool to perform a multivariate goodness-of-fit (GoF) test originally developed to discover new physics phenomena in experimental collider data \cite{DAgnolo:2018cun,Letizia:2022xbe,Grosso:2023scl}. Inspired by the Neyman--Pearson framework, NPLM estimates the ratio between two samples' densities to construct a likelihood-ratio test.  
We demonstrate how NPLM can be deployed to assess the goodness of generative models, particularly in high-dimensions. In this work, we review the main principles of the algorithm and show how its output can be used to quantitatively compare different generative models, evaluate the quality of a given model across the data support, and provide guidance for generation improvement. Here, we focus on the implementation of NPLM based on kernel methods proposed in \cite{Letizia:2022xbe}. For a comparison between different testing metrics and the NPLM, we refer the reader to Ref.~\cite{Grossi:2025pmm}.

This work is organized as follows. In Section \ref{sec:NPLM}, we review the practice of two-sample testing for goodness-of-fit to assess the compatibility between a given reference distribution, $p_R$, and a set of data from a distribution $q$, and then describe the NPLM approach. In Section \ref{sec:Experiments}, we present our results for two cases: the performance of normalizing flows (NFs) trained on mixtures of Gaussians with increasing dimensionality, and the performance of \textit{FlowSim} \cite{vaselli2024endtoend}, an NF-based model trained on simulated LHC-like jet data. In Section \ref{sec:Diagnosis}  we showcase how the method can be used as a diagnostic tool to identify mismodelled regions in the data. Finally, in Section \ref{sec:Conclusions}, we present our conclusions and outlook.

%% file: Methodology.tex
The goal of a GoF  test is to evaluate the compatibility between a reference statistical model $p_R$ and a dataset of interest sampled from a distribution $q$. If the former is not available in closed analytical form but it can be sampled, for example via Monte Carlo simulations, the test can be designed as a two-sample test (2ST). To this end, let us  consider a set $\scriptR=\{x_1,...,x_{N_\scriptR}\}$ drawn from the reference distribution (a reference sample),
and denote the data sample as $\scriptD=\{\tilde{x}_1,...,\tilde{x}_{N_\scriptD}\}$, where $x_i, \tilde{x}_i\in\mathcal{X}\subseteq\Reals^d$. Let us define the null hypothesis $H_0$ of the test as the hypothesis under which both the reference and the data samples are drawn from the reference distribution $p_R$.
A typical two-sample test proceeds by selecting a \emph{test statistic}, namely a real-valued function 
\be 
t:\mathcal{X}^{N_\scriptR}\times\mathcal{X}^{N_\scriptD}\rightarrow\Reals.
\ee
In a test of hypotheses, a large value of the test statistic indicates a potential tension with the null hypothesis. To quantify this statement, the distribution of the test under the null hypothesis $p(t|H_0)$ needs to be known or estimated. Consequently, the $\textit{p-value}$ is defined as
\be\label{eq:p-value}
\textrm{p}_{\rm value}=P(t\geq t_{\rm obs}|H_0)=\int_{t_{\rm obs}}^\infty p(t|H_0) dt,
\ee 
and it represents the probability, under the null hypothesis, to obtain data as extreme as the observed ones. The result of the test is considered statistically significant if the observed p-value is smaller than a pre-selected rate of type-I errors (false positive rate) 
\be\label{eq:fpr}
\alpha = P(t\geq t_\alpha|H_0).
\ee
The statistical interpretation can also be expressed in terms of the $\textit{Z-score}$ defined as

\be\label{eq:Zscore}
\mathrm{Z_{score}} = \Phi^{-1}(\mathrm{1-p_{value}})
\ee
where $\Phi^{-1}$ is the inverse cumulative distribution function of the standard normal. In this case, a larger Z-score would indicate a greater deviation of $\scriptD$ from $p_R$.

As anticipated earlier, we are interested in leveraging this framework to perform a GoF test. Accordingly, we estimate the distribution of the test statistic under the null hypothesis by testing the reference distribution against itself. In practice, we do this by evaluating the test statistic on a reference sample $\scriptR$ and a number $N_{\rm toys}$ of independent reference-distributed samples $\scriptD_i^{(R)}$, that are sometime referred to as \emph{pseudo-experiments}, \emph{toy datasets} or simply \emph{toys}. Each of these tests returns a value $t_i=t(\scriptD_i^{(R)},\scriptR)$. The set $\{t_i\}_{i=1}^{N_{\rm toys}}$ can be used to empirically estimate the p-value as
\be\label{emp_pvalue}
\hat{\textrm{p}}_{\rm value} = \frac{1}{N_{\rm toys}+1} \left[\sum_{i=1}^{N_{\rm toys}} \mathds{1}(t_i-t_{\rm obs})+1\right],
\ee
where $\mathds{1}(x)$ is the Heaviside step function, which is zero when $x<0$ and one otherwise. However, given that modern generative models have yet to reach a high degree of fidelity set by the standards of HEP research, the typical p-values that we will observe in this study are small and would require many pseudo-experiments to be estimated precisely. For this reason we rely on approximating the null distribution $p(t|H_0)$ with a $\chi^2$ distribution with a number of degrees of freedom that is selected with a fit to the values $\{t_i\}_{i=1}^{N_{\rm toys}}$ of the test statistic under the null hypothesis. This approximation is good at an empirical level and we check that this is the case for every scenario we investigate. However, at this point, we do not have a clear understanding on why this compatibility is observed (see \cite{DAgnolo:2018cun, Letizia:2022xbe, Grosso:2023scl} for further details). This aspect and potential connection with Wilk's theorem \cite{10.1214/aoms/1177732360} will be subject of future theoretical studies. Finally, we would like the effect of the statistical fluctuations of the reference sample $\scriptR$ on the outcome of the test to be subdominant. To this end, we consider an unbalanced setup with $N_\scriptR \gg N_\scriptD$.

As anticipated earlier, we aim at assessing the compatibility between a candidate generative model and the true distribution of the data. To this end, we consider two setups. In the first one, the true distribution of the data is promoted to a reference distribution and we assess its compatibility with respect to a set of data sampled from a generative model $p_G$. In the second case, we consider the generative model as the reference and assess its capability to generate data that are distributed according to the true distribution.

In real-world applications, the number of available reference events is often limited, for instance owing to the computational cost of Monte Carlo simulations. This makes it difficult to estimate the distribution of the test statistic as outlined earlier. However, it is possible to address this scarcity of data by leveraging re-sampling techniques, with or without replacement.
Finally, it is important to perform the GoF test on a test set of generated data that is independent of the training set use to optimize the generative model.

\begin{figure}[t!]
    \centering
    \includegraphics[width=0.8\textwidth]{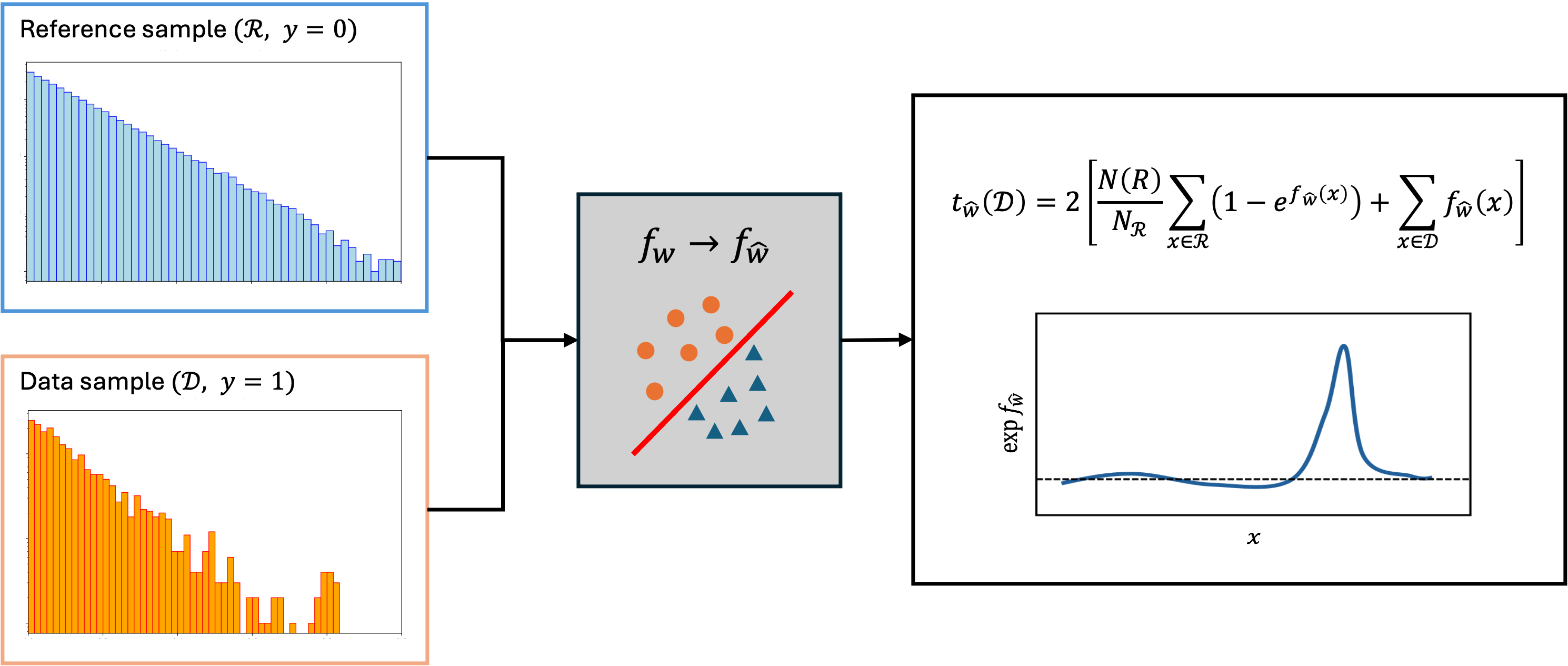}
    \caption{Schematic representation of the NPLM test.}
    \label{fig:nplm_scheme}
\end{figure}

\subsection{The New Physics Learning Machine}\label{subsec:NPLM}
The NPLM test is based on the idea of modeling the probability density of the data of interest $q$ as a local deformation of the reference density
\be\label{nplm_deformation}
q_w(x)=e^{f_w(x)}p_R(x),
\ee
with $\mathcal{F}=\{f_w\}$ a rich family of functions parametrised by $w$. Note that the densities in Eq.~\eqref{nplm_deformation} can be in general normalized to different total masses, although this is not relevant for our specific use case. In \cite{Letizia:2022xbe} and in this work, we consider kernel methods based on the Gaussian kernel $k_\sigma (x,x')=\exp \left(-\lVert x-x'\lVert^2/2\sigma^2\right)$, with $\sigma$ the kernel width.  The function $f_w$ is then expressed as a weighted sum
\be\label{kernel_methods}
f_w(x)=\sum_{i=1}^{N} w_i k_\sigma (x,x_i),
\ee
with $N=N_\scriptR+N_\scriptD$. Concretely, the optimal parameters $\hat{w}$ are learned from data with a supervised classifier trained on the reference data $\scriptR$ and the data sample $\scriptD$. The model is trained to minimise a regularized empirical risk of the following form
\be\label{reg_ERM}
\hat{L}(w)=\frac{1}{N}\sum_i\ell (y_i,f_w(x_i))+\lambda R(f_w),
\ee
where the first term is a weighted logistic loss
\be
\ell(y,f_w(x))=(1-y)\frac{N(R)}{N_\scriptR} \log\left(1+e^{f_w (x)}\right)+y\log\left(1+e^{-f_w (x)}\right),
\ee
with $y=0$ if $x\in\scriptR$ and $y=1$ if $x\in\scriptD$. In this specific application, the number of points $N_\scriptD$  in the data sample is not the realisation of a random variable as in the case of a typical HEP data analysis. Hence, we simply identify the number of expected events $N(R)$ with $N_\scriptD$. This loss can be shown (see \cite{Letizia:2022xbe}) to have the correct target function
\be\label{eq:f}
f_{\hat{w}}(x)\approx f^*(x)=\argmin_{f}\mathds{E}\left[\ell(y,f(x))\right] = \log \frac{q(x)}{p_R(x)}.
\ee
The second term in Eq.~\eqref{reg_ERM} is a regularisation term
\be
R(f_w)=\sum_{ij} w_i w_j k_{\sigma}(x_i,x_j),
\ee
 constraining the complexity of the model, weighted by the regularization parameter $\lambda$. At the end of training, the model is evaluated in-sample on the whole training data with the following metric
\be
t_{\hat{w}}(\scriptD,\scriptR)=-2\left[\frac{N(R)}{N_\scriptR}\sum_{x\in\scriptR}\left(e^{f_{\hat{w}}(x)}-1\right)-\sum_{x\in\scriptD} f_{\hat{w}}(x)\right],
\ee
which is derived from the extended likelihood-ratio (see \cite{barlow1990extended,DAgnolo:2018cun,Letizia:2022xbe}). This method allows to apply the Neyman--Pearson approach to hypothesis testing for GoF testing by relying on a data driven alternative hypothesis. Beside the value of the test statistic with its p-value, the function $f_{\hat{w}}$ learned by the model can be used to extract additional information on the potential discrepancy between the reference distribution $p_R$ and the data of interest. For instance, the function is expected to evaluate to approximately zero where the two distribution are close and be different from zero where they differ. This can be inspected directly by re-weighting a set of reference data. Similarly, being NPLM based on classifiers, an anomaly score can be assigned to individual data point for further analysis. We will reprise these topics in the following section. A representation of the NPLM method is depicted in Figure~\ref{fig:nplm_scheme}. We refer the reader to Appendix \ref{app:modelselection} for the model selection pipeline and Ref.s~\cite{Letizia:2022xbe, Grosso:2023scl} for further details.

%% file: MoG_NFs.tex
In this section, we demonstrate how to apply the NPLM method to the problem of validating and evaluating generative models. This work focuses particularly on Normalising Flows (NFs)~\cite{rezende2015variational,kobyzev2020normalizing}. These models are widely adopted in scientific applications because they provide explicit likelihoods that can be both sampled and evaluated. Nevertheless, this approach can be applied to any generative model since the analysis methods are not specific to NFs.

As outlined in Section~\ref{sec:NPLM}, we perform two complementary sets of experiments:
\begin{itemize}
    \item{\bf True data distribution as reference.} We fix the true distribution of the data as the reference distribution and test samples from the generative model $p_G$ against it. This approach has the advantage that the null hypothesis, which depends only on samples from the reference, is universal. Once estimated, we can use it to test any generative model designed to model the true distribution. However, we may have limited reference data, possibly requiring resampling methods.
    \item{\bf Generative model as reference.} We treat the generative model as the reference and test whether it is a good fit to samples drawn from the true distribution. The advantage here is that the generative model can produce arbitrarily many samples efficiently, but the null hypothesis must be estimated anew for each model we wish to test.
\end{itemize}
These two approaches will generally yield different results when the true distribution and the generative model exhibit detectable discrepancies.
\subsection{Mixture of Gaussians with Normalizing Flows}\label{sec:MoG}

As a first use case, we consider a synthetic benchmark in which we have full control of the data-generating distributions.

\paragraph{Dataset}
 Following refs. \cite{Coccaro:2023vtb,Reyes-Gonzalez:2022rco}, we consider a set of mixtures of Gaussians (MoG) with $n_{\rm comp}=3$ components in $d=4,8,20,30$ dimensions. Each component is characterized by randomly generated means and standard deviations in the $[0,10]$ and $[0,1]$ intervals, respectively. They are then mixed according to a $n_{\rm comp}$-dimensional categorical distribution with randomly generated probabilities. The resulting multivariate distributions have random off-diagonal elements of order one in the covariance matrix and multi-modal marginals. For each distribution, we trained three NF models characterized by 5 RealNVP bijectors \cite{dinh2016density}, made of 3 layers of 64 nodes each. They have been optimized using $\text{N}_{\text{tr}}=10^5$, $2\cdot 10^5$ and $5\cdot 10^5$ training samples.

\begin{table}[t!]
\centering
\begin{tabular}{|c|cccc|}
\hline
\diagbox{$\text{N}_{\text{tr}}$}{$d$}&
\multicolumn{1}{c|}{4} & \multicolumn{1}{c|}{8} & \multicolumn{1}{c|}{20} & 30 \\ \hline
\multicolumn{1}{|c|}{100k}      & \multicolumn{1}{c|}{$9.88 \> ^{+1.22}_{-1.29}$}   & \multicolumn{1}{c|}{$8.88 \> ^{+1.12}_{-1.19}$}    & \multicolumn{1}{c|}{$14.46 \> ^{+1.09}_{-0.84}$}    & {$14.97 \> ^{+1.09}_{-0.84}$}    \\ \hline
\multicolumn{1}{|c|}{200k}      & \multicolumn{1}{c|}{$4.79 \> ^{+1.00}_{-1.07}$}   & \multicolumn{1}{c|}{$9.90 \> ^{+0.94}_{-1.05}$}    & \multicolumn{1}{c|}{$6.45 \> ^{+0.97}_{-1.07}$}    &  {$7.32 \>^{+0.90}_{-0.81}$}   \\ \hline
\multicolumn{1}{|c|}{500k}      & \multicolumn{1}{c|}{$1.93 \> ^{+1.02}_{-0.99}$}   & \multicolumn{1}{c|}{$ 3.01\> ^{+0.74}_{-1.13}$}    & \multicolumn{1}{c|}{$2.07 \> ^{+0.81}_{-0.97}$}    &   {$3.06 \> ^{+1.13}_{-0.86}$}  \\ \hline
\end{tabular}
\caption{Table of median Z-scores obtained with the NPLM method for various NFs models, characterized by training samples of different size (${\rm N}_\text{tr}$) and different number of dimensions ($d$).}\label{tab:Z_scores_NF_fullsim}
\end{table}

\begin{table}[h]
\centering
\begin{tabular}{|c|cccc|}
\hline
\diagbox{$\text{N}_{\text{tr}}$}{$d$}&
\multicolumn{1}{c|}{4} & \multicolumn{1}{c|}{8} & \multicolumn{1}{c|}{20} & 30 \\ \hline
\multicolumn{1}{|c|}{100k}      & \multicolumn{1}{c|}{$7.31 \> ^{+0.81}_{-0.95}$}   & \multicolumn{1}{c|}{$ 6.77\> ^{+0.97}_{-0.85}$}    & \multicolumn{1}{c|}{$10.20\> ^{+1.17}_{-0.86}$}    & {$9.75 \> ^{+0.80}_{-0.94}$}    \\ \hline
\multicolumn{1}{|c|}{200k}      & \multicolumn{1}{c|}{$3.03 \> ^{+0.96}_{-1.17}$}   & \multicolumn{1}{c|}{$6.32 \> ^{+0.90}_{-0.82}$}    & \multicolumn{1}{c|}{$4.05 \> ^{+1.11}_{-0.84}$}    &  {$4.88 \> ^{+0.94}_{-0.65}$}   \\ \hline
\multicolumn{1}{|c|}{500k}      & \multicolumn{1}{c|}{$1.02 \>^{+0.94}_{-0.99}$ }   & \multicolumn{1}{c|}{$1.12 ^{+1.04}_{-0.78}$}    & \multicolumn{1}{c|}{$0.87 ^{+0.81}_{-0.7}$}    &   {$1.78 ^{+0.82}_{-0.77}$}  \\ \hline
\end{tabular}
\caption{Inverted hypotheses. Table of median Z-scores obtained with the NPLM method for various NFs models, characterized by training samples of different size (${\rm N}_\text{tr}$) and different number of dimensions ($d$).}\label{tab:Z_scores_NF_gen}
\end{table}

\begin{table}[t!]
    \centering
    \begin{tabular}{|c|c|c|}
    \hline
    $N_\scriptD$ & MoG as ref. & NF as ref. \\ \hline\hline
    5 k               &   $0.39^{+0.95}_{-0.64}$   &     $0.38^{+0.86}_{-0.92}$         \\ \hline
    10 k              &   $1.93^{+1.02}_{-0.99}$   &    $1.02^{+0.94}_{-0.99}$        \\ \hline
    20 k              &   $3.70^{+1.18}_{-1.29}$   &    $1.93^{+1.16}_{-0.85}$      \\ \hline
    40 k              &   $5.93^{+1.56}_{-1.49}$   &    $4.05^{+1.02}_{-1.04}$        \\ \hline
    \end{tabular}
    \caption{Z-scores of the NPLM test for data generated by a NF  in 4D with $\text{N}_{\text{tr}}=500k$ at varying sample size.  We report a $68\%$ confidence interval.}\label{tab:NF-varyN}
\end{table}
\begin{figure}[t!]
    \centering
    \includegraphics[width=0.7\linewidth]{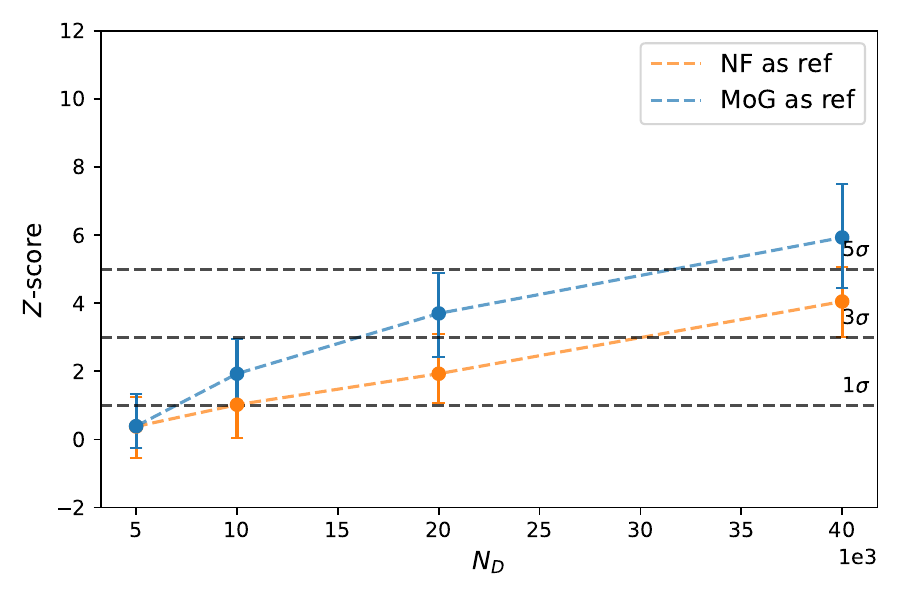}
    \caption{Validation Z-scores of the NF, trained on the 4D-MoG with 500k samples, as function of the sample size used for the test. The blue line represents MoG as the reference case, while NF as the reference case is shown in orange. }
    \label{fig:MoG_Zscore}
\end{figure}

\paragraph{NPLM test}

\begin{figure}[t!]
    \centering
    \includegraphics[width=0.7\textwidth]{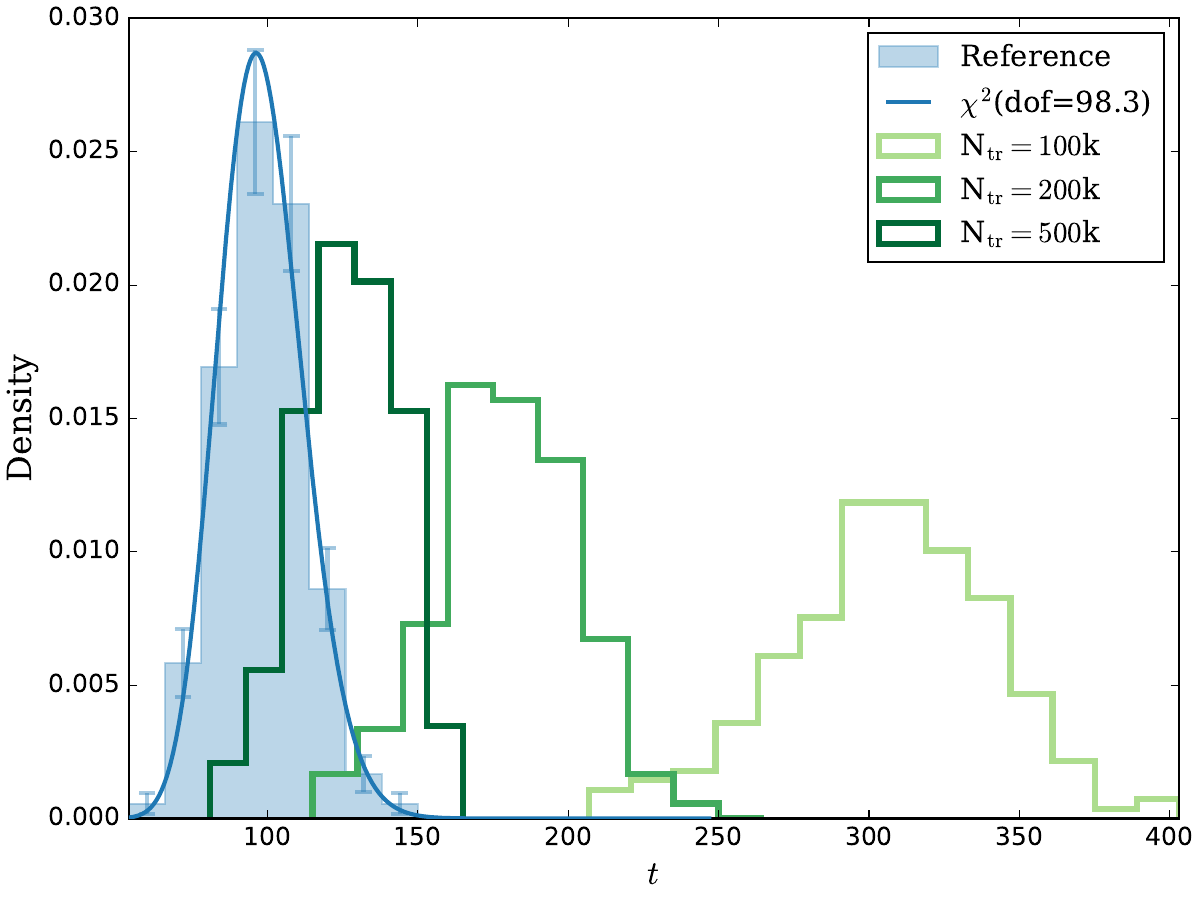}
    \caption{Empirical distribution of the NPLM test ($t$) for NF models with $D=4$ and $\text{N}_{\text{tr}}=100k,\,200k,\,500k$. (shades of green). The null empirical distribution represented by the light blue histogram is compared with a $\chi^2$ distribution with $98.3$ degrees of freedom.}
    \label{fig:MOG_rank_4D}
\end{figure}

We select the hyperparameters of the kernel-based NPLM model according to the prescription given in Refs.~\cite{Letizia:2022xbe} and summarised in Appendix~\ref{app:modelselection}.
The resulting best values for the three hyperparameters are: $M=10^4$ and $\lambda = 10^{-10}$, while $\sigma$ depends on the MoG dimension $d$ and is given by the following values $\sigma= 4.96,6.08,9.69,10.9$ for $d=4,8,20,30$ respectively. The test is performed with $N_\scriptR=10^5$ reference points. The number of data points $N_\scriptD$ varies depending on the scope of the experiment.
\paragraph{Results}
Tables \ref{tab:Z_scores_NF_fullsim} and \ref{tab:Z_scores_NF_gen} show the median Z-scores resulting from the NPLM test. The first table is obtained by fixing the true distribution of the data as the reference, while in the second the generative model is the reference.
The errors reported in the table represent the 68\% confidence interval around the median.
Fig. \ref{fig:MoG_Zscore}  show a systematic trend between the magnitude of the Z-score metric and the size of the training set, i.e. the larger is the training sample the more accurate the generative model is, resulting in a lower Z-score. This trend is also shown in Figure~\ref{fig:MOG_rank_4D}, where the empirical distributions of the NPLM tests for the 4-dimensional case, for the null and alternative hypotheses. 
We observe stronger discrepancies when the true distribution is chosen as the reference. This can be attributed to the asymmetry between the true and generative supports: a generative model often assigns non-negligible probability mass to regions where the true data have little or none, making deviations more visible when testing model samples against true data. Conversely, when the generative model is the reference, its broader support tends to accommodate true samples more easily, leading to weaker apparent tension.
Table \ref{tab:NF-varyN} and Fig. \ref{fig:MoG_Zscore} show the results obtained when the sample size of the data sample is varied. Samples generated with size $N_{\scriptD}<1.5 \cdot 10^4$ and $N_{\scriptD}<7\cdot 10^3$ show a median discrepancy not greater than $3\sigma$ and $1\sigma$ respectively, when the MoG model is the reference. When the NF is the reference, we obtain $N_\scriptD<3 \cdot 10^4$ and $N_\scriptD<10^4$.

%% file: FlashSim.tex
\subsection{Particle Jets with FlowSim}\label{sec:FlashSim}

As a second use case, we study the generative model proposed in Ref.~\cite{vaselli2024endtoend}. This model, dubbed FlowSim, is a Normalizing Flow– and Flow Matching–based approach to fast simulation of high-energy collision events~\cite{lipman2023flowmatchinggenerativemodeling}. FlowSim delivers simulation quality in good agreement with traditional Monte Carlo–based methods while achieving orders-of-magnitude speed-ups. In Ref.~\cite{vaselli2024endtoend}, the model is trained not on full \textsc{geant4}-level simulations \cite{GEANT4:2002zbu}, which would be the realistic target, but on a simplified collider dataset: events are generated with \textsc{pythia8} \cite{Bierlich:2022pfr} and passed through a lightweight, \emph{Delphes}-like \cite{deFavereau:2013fsa} toy detector that introduces smearing, flavour-dependent responses, and realistic correlations among jet kinematics, tagging variables, and constituent properties. This setup provides a controlled yet detector-like environment, with full details available in the original paper. Our objective is to evaluate FlowSim by comparing it to the traditional Monte Carlo–based simulation pipeline, referred to as FullSim, which serves as the standard for current LHC analyses and the target for generative model replication.

\paragraph{Dataset} 
FlowSim aims at simulating jets, namely sprays of particles originating in proton-proton collisions observed in particle physics experiments at the LHC.\footnote{See \cite{vaselli2024endtoend} for a detailed description of the physics problem and properties of jets.} Each data point in the dataset represents a jet object generated from a t\={t} process at LHC, and is described by 16 observables. The features describing each jet are divided in basic reconstructed, namely $p_T$, $\eta$, $\phi$, invariant mass, b-tagging discriminator and number of constituents; and extended dataset variables, namely neutral hadron fraction (nhf), charged hadron fraction (chf), neutral electromagnetic fraction (nef), charged electromagnetic fraction (cef), quark-gluon discriminator (qgd), jet identification (JetId), number of charged and neutral particles, c-tagging discriminator and number of secondary vertices (nSV). An in-depth description of these variables and how they are generated or reconstructed by a generative model can be found in Ref.~\cite{vaselli2024endtoend}.

\paragraph{Task}

The testing procedure follows the methodology described in Section~\ref{sec:MoG}. 
As in the previous case, two tests are performed depending on the chosen reference model. 
The reference sample size is fixed to 1 million events, while the data sample size varies between 5{,}000 and 200{,}000 events. 
The model hyperparameters are determined according to the strategy outlined in Section~\ref{subsec:MoG}, consistent with the previous benchmark, yielding $M = 8000$,  $\lambda = 10^{-6}$, and $\sigma = 7.4$. Once the NPLM model is defined, a distinct null hypothesis is estimated for each sample size. 
Each estimation is based on $N_{\rm toys} = 200$ toy datasets. 
The distribution of the test statistic under the alternative hypothesis is obtained by repeating the test 40 times. 
Given the finite availability of FullSim data ($\approx$ 15 million events), each training is performed on a bootstrapped sample drawn from the complete FullSim dataset. 
Similarly, for the FlowSim model, we rely on bootstrapping from a total of approximately 4 million sampled events for the analysis.

\begin{figure}[t!]
    \centering
    \includegraphics[width=0.7\linewidth]{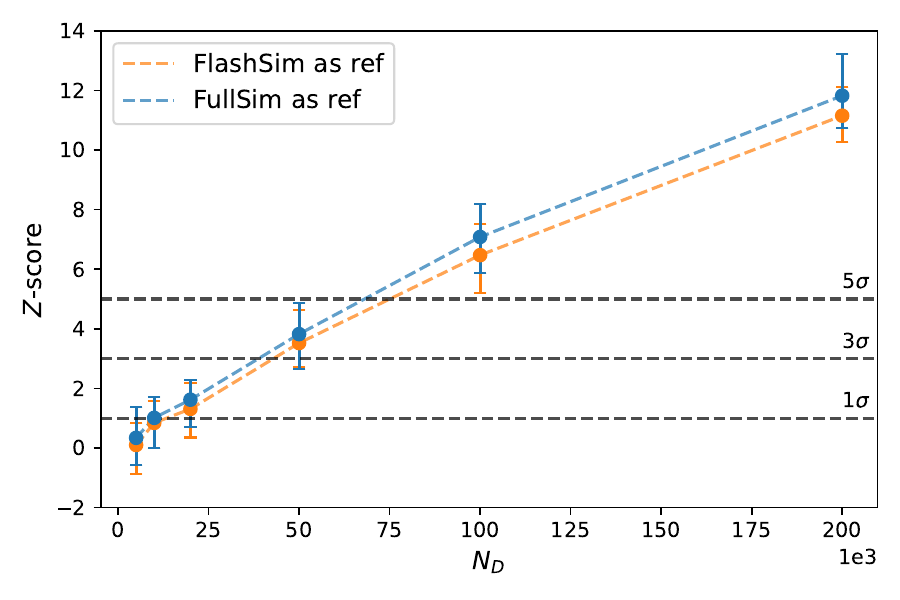}
    \caption{Z-scores of the FlowSim validation as function of the size of the analyzed FlowSim dataset. The blue line represents FullSim as the reference case, while FlowSim as the reference case is shown in orange. }
    \label{fig:FlashSim_Zscore_vs_ND}
\end{figure}

\begin{table}[t!]
    \centering
    \begin{tabular}{|c|c|c|}
    \hline
    $N_\scriptD$ & FullSim as ref & FlowSim as ref \\ \hline\hline
    5 k               &   $0.34^{+1.03}_{-0.92}$  & $0.1^{+0.73}_{-0.96}$             \\ \hline
    10 k              &   $1.01^{+0.71}_{-1.02}$   & $0.84^{+0.75}_{-0.85}$              \\ \hline
    20 k              &   $1.62^{+0.66}_{-0.92}$   & $1.31^{+0.88}_{-0.96}$               \\ \hline
    50 k              &   $3.82^{+1.05}_{-1.18}$   &  
    $3.52^{+1.12}_{-0.8}$  
        \\ \hline
    100 k             &   $7.08^{+1.10}_{-1.22}$   &   
    $6.47^{+1.04}_{-1.26}$
        \\ \hline
    200 k             &   $11.82^{+1.41}_{-1.09}$  &    
    $11.15^{+0.96}_{-0.89}$
        \\ \hline
    \end{tabular}
    \caption{Z-scores of the NPLM test for FlowSim data at varying sample size.}\label{tab:FlashSim-results}
\end{table}

\paragraph{Results}
Table \ref{tab:FlashSim-results} reports the resulting Z-scores for different sample sizes. As in the previous section, the error associated to the Z-score represents the $68\%$ confidence interval around the median. In Fig. \ref{fig:FlashSim_Zscore_vs_ND}, we observe that the level of discrepancy between the reference model and the data increases with the sample size $N_\scriptD$. This is expected and it is a further indication of the consistency of our testing methodology. As before, we find that using the generative model as the reference leads to lower median Z-scores. Nonetheless, the observed difference between the reference choices is now within the uncertainties. Furthermore, we estimate that in order to obtain a discrepancy lower than $1\sigma$, $3\sigma$, and $5\sigma$, the sample sizes must be approximately $10^4$, $4 \cdot 10^4$, and $7 \cdot 10^4$, respectively, when FullSim is the reference, and $10^4$, $5 \cdot 10^4$, and $8 \cdot 10^4$, respectively, when FlowSim is the reference.

%% file: Diagnosis.tex
Extracting knowledge about the nature and location of a detected discrepancies is crucial to evaluate their impact in presence of uncertainties. Here we showcase how to use the NPLM model as a diagnostic tool to gain insights about the most anomalous aspects of the generated data.

\begin{figure}[t!]
    \centering
    \includegraphics[width=0.49\linewidth]{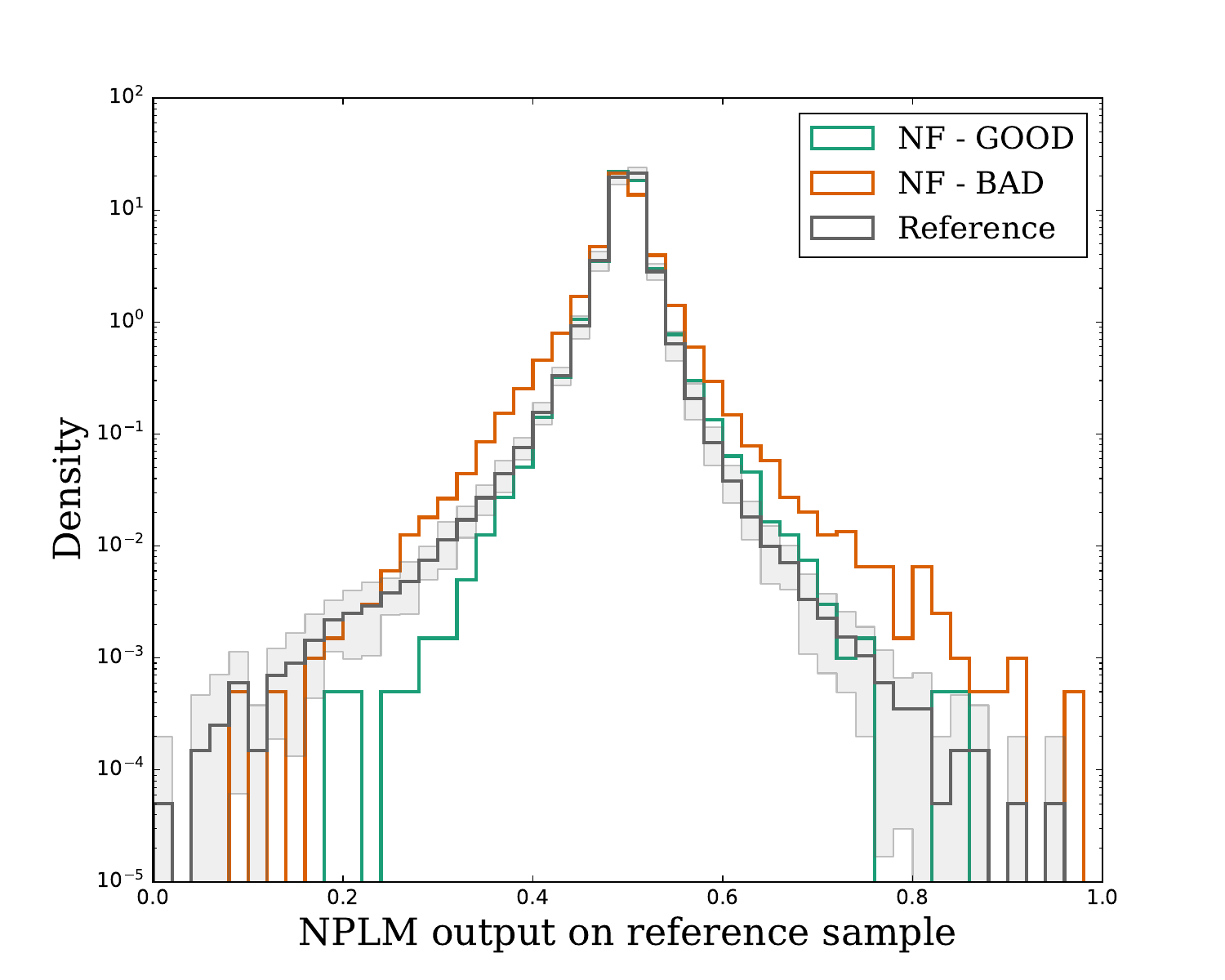}
    \includegraphics[width=0.49\linewidth]{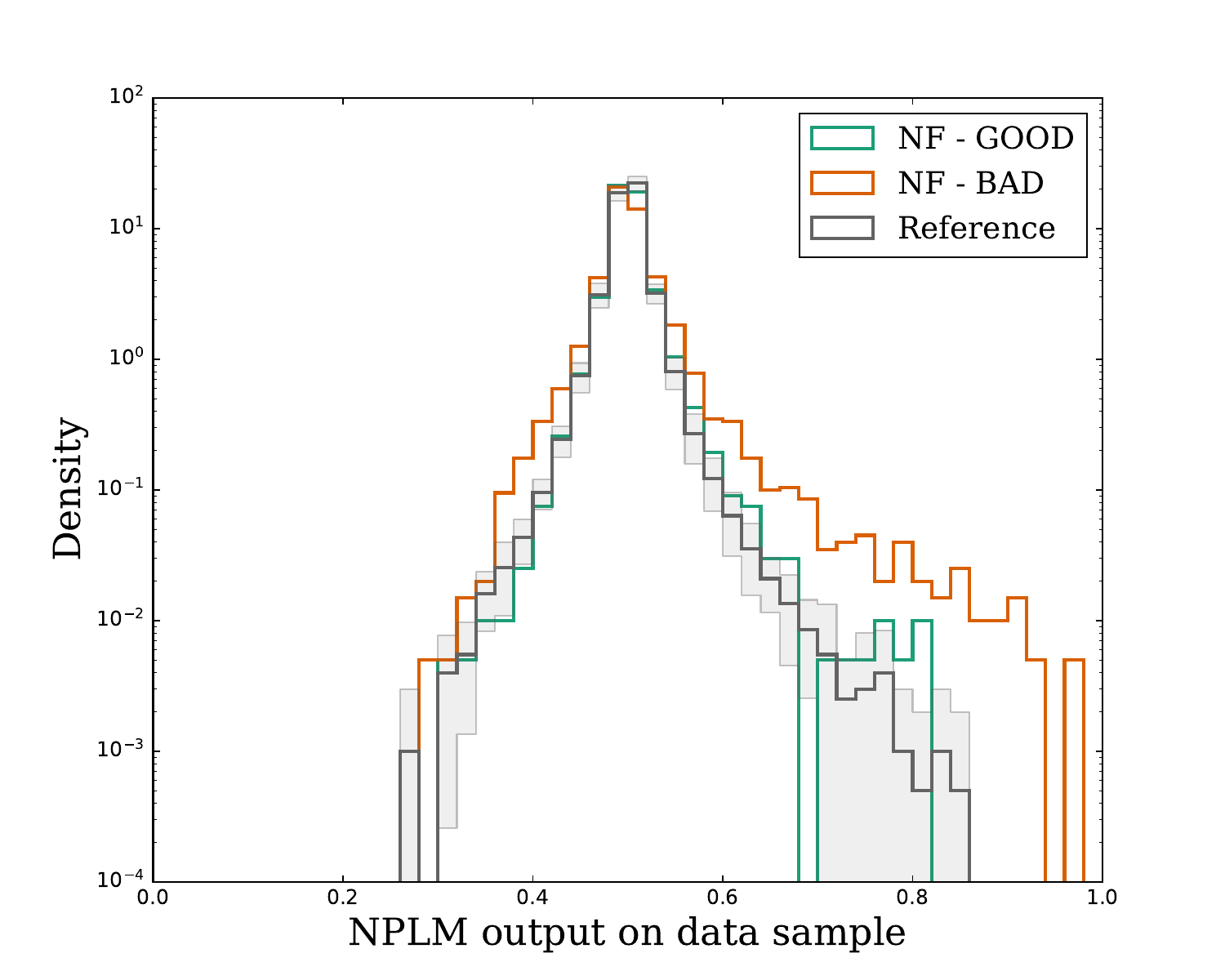}
    \caption{Examples of the NPLM classifier score distributions evaluated on reference data and data for the NF experiments. The green and orange histograms are the output of the model trained on data from the most accurate and the least accurate NFs respectively. The grey histogram represents the mean over ten reference-distributed toys. The region in light grey covers one standard deviation around the mean.}
    \label{fig:scores}
\end{figure}

\begin{figure}[t!]
    \centering
    \includegraphics[width=0.49\linewidth]{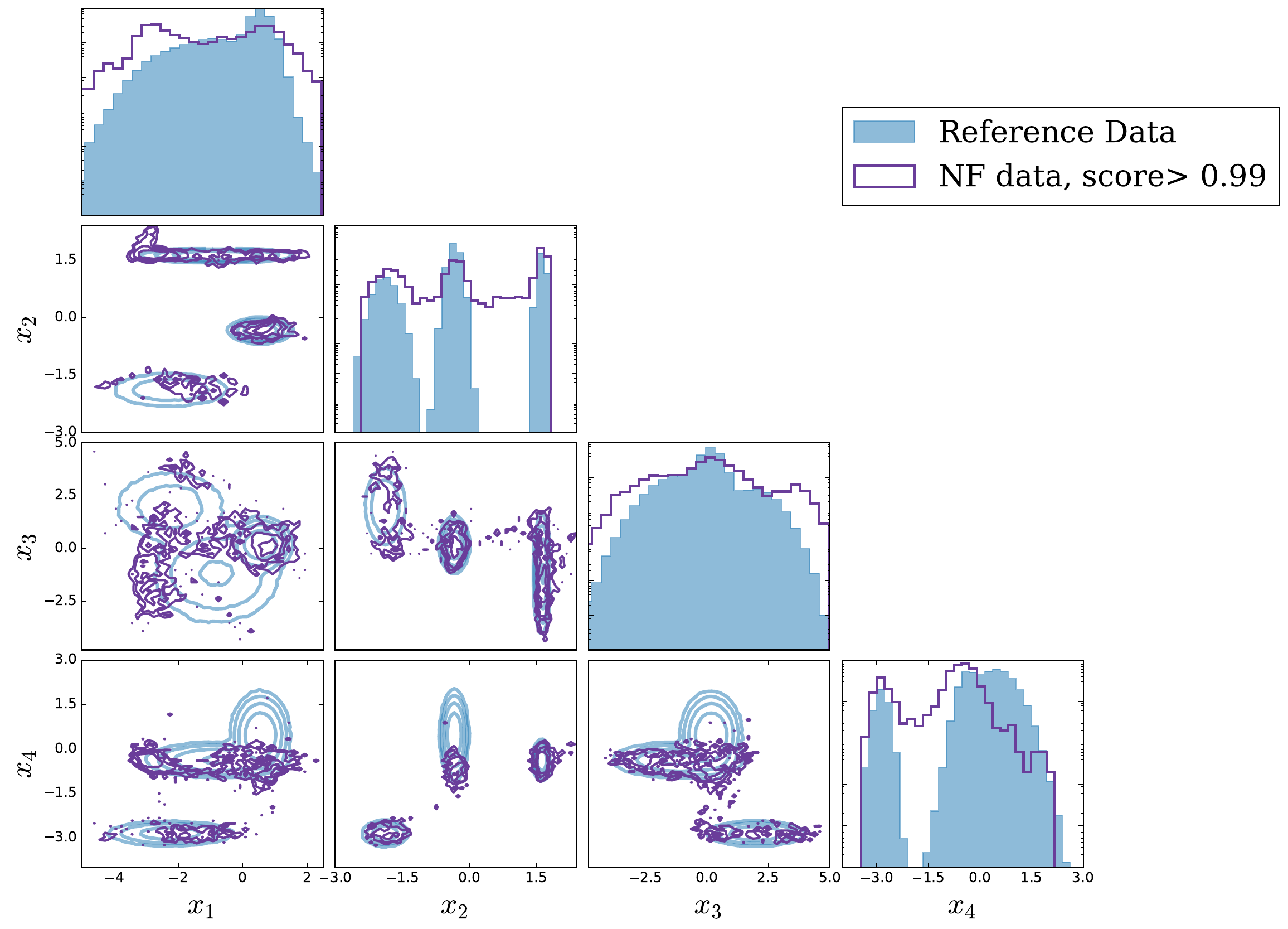}
    \includegraphics[width=0.49\linewidth]{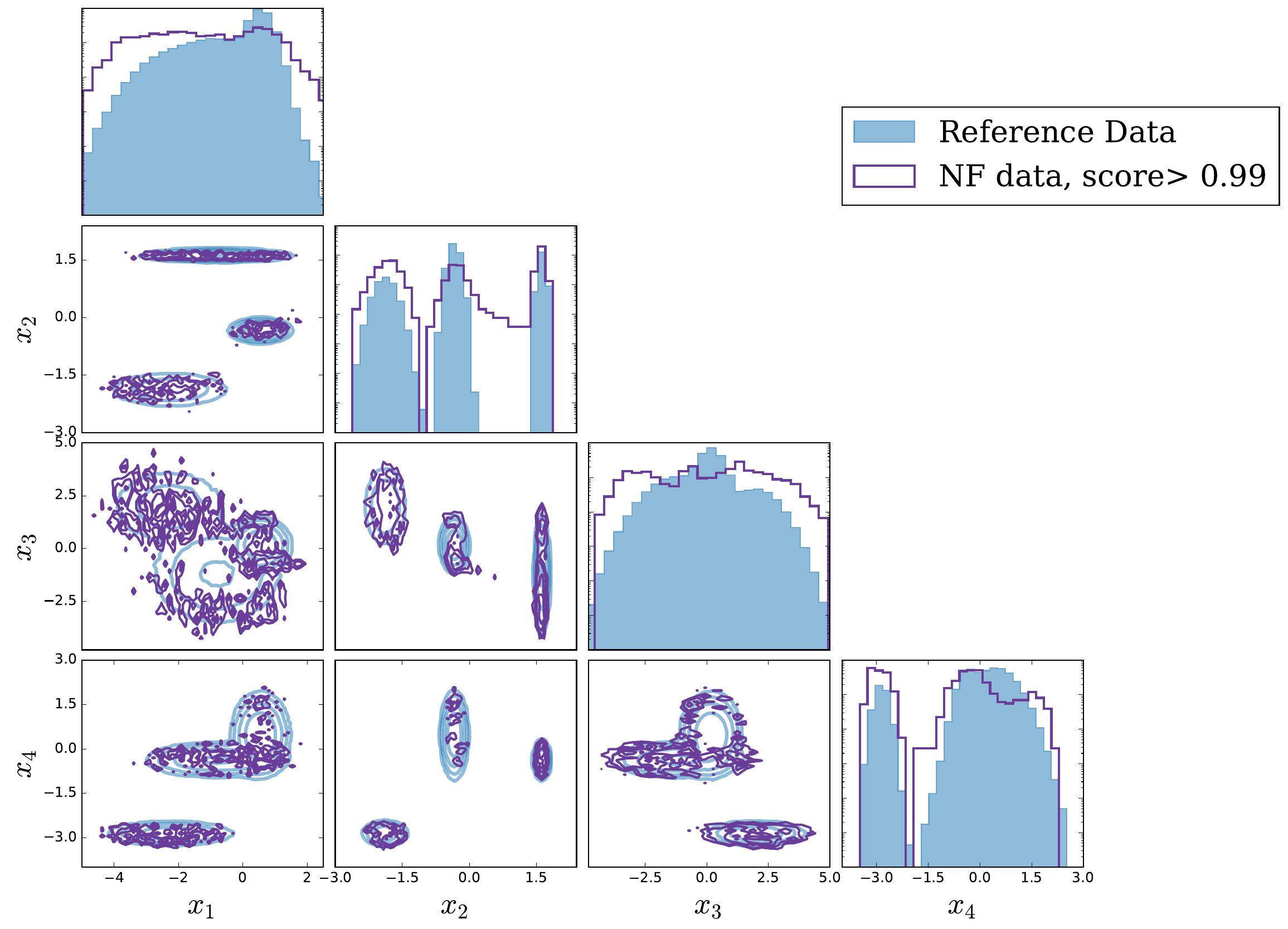}
    \caption{Anomaly score distributions. In purple, the top 1\% scored generated data points are shown, while the reference data is depicted in blue for the bad (left panel) and the good (right panel) NF model.}
    \label{fig:MoG-anomaly}
\end{figure}
As an illustrative example throughout this paragraph, we focus on two of the 4-dimensional NF models, with the true distribution as the reference, trained with $\text{N}_\text{tr}=100k$ and $\text{N}_\text{tr}=500k$, for which the NPLM test reports a median Z-score of $9.88$ and $1.93$ respectively.
The two models provide examples of poor and good modeling.

We can take advantage of the underlying classifier model and look at its score, defined as the sigmoid transformation of the function $f_{\hat{w}}(x)$ that the NPLM model has learnt. This allows us to inspect the local nature of the discrepancy and assign a score to each data point. In presence of discrepancies, the data points laying in the discrepant regions are assigned a more extreme score, closer to one for regions of overdensity and to zero for underdensities.
Figure~\ref{fig:scores} shows two examples of score distributions computed on the reference data (left panel) and the data (right panel). The green and orange histograms show the model outputs trained on data from the most and least accurate NFs, respectively. The gray histogram indicates the mean across ten reference-distributed samples, while the light gray band represents one standard deviation around that mean. The gray histograms are approximately distributed around $0.5$ with a certain variance and a relatively small excess of points in the right-hand tail. This is due to the fact that NPLM tends, by construction, to slightly overfit the training data. We observe that both NF models have a more pronounced right tail due to mismodeling causing overdensities, especially the least accurate model. In both case there is no strong evidence for underdensities. One can apply a selection of the data points based on the score to study the properties of the anomalies via diagnostic plots. 

The NPLM framework allows us to investigate mismodelling on an event-by-event basis. Furthermore, it enables us to explore not only the marginal variables but also their correlations. Figure~\ref{fig:MoG-anomaly} compares the distribution of the top 1\% scored generated data points with the reference sample via corner plots for the two NF models. This plots inform us about the nature and location of mismodelled marginals and correlations. In the specific examples showed in the figures, we can clearly see that the worst model (left panel) has a hard time reproducing correctly the distribution of data points in the tails and their correlations. We can also see how the outcome improves with the better model (right panel), where the distribution of mismodelled points is more uniform since the high anomalous scores are mostly driven by statistical fluctuations.

%% file: nplm.tex
To increase the efficiency of the model while preserving performance, we rely on Falkon \cite{meanti2020kernel}, a modern solver for large-scale kernel methods. The basic idea is to replace Eq.~\eqref{kernel_methods} with
\be\label{nystrom_km}
f_w(x)=\sum_{i=1}^{M} w_i k_\sigma (x,x_i),
\ee
where $\{\tilde{x}_1,...,\tilde{x}_M\}$ are called Nyström centres and are sampled uniformly at random from the input data, with $M$ an hyperparameter. The corresponding solution can be shown to be with high probability as accurate as the exact one (see \cite{rudi2017generalization} and references therein). Problem~\eqref{reg_ERM} is then solved  by an approximate Newton iteration, as discussed in details in \cite{meanti2020kernel}. 
Ultimately, the Falkon-based implementation of NPLM is characterised by three main hyperparameters: the number of centres $M$, the kernel width $\sigma$ and the regularisation parameter $\lambda$. These are tuned only on reference data to avoid biases toward specific anomalous features that might be present in the data of interest $\scriptD$. Following \cite{Letizia:2022xbe,Grosso:2023ltd}, the criteria for their selection can be summarised as:
\begin{itemize}
\item The Gaussian width $\sigma$ is selected as the 90th percentile of the pairwise distance among reference-distributed data points. Heuristics of this type are commonly used for kernel methods, see for instance \cite{gretton2012kernel}.
\item To achieve optimal statistical bounds and preserve performance, the number of centres $M$ must be at least be of order $\sqrt{N}$, as discussed in~\cite{rudi2016more}.
Studies presented in \cite{Letizia:2022xbe} suggest that values close to the number of data points  ${N}_{\cal{D}}$ in the measurements work well but can be reduced for a faster training.
\item The regularisation parameter $\lambda$ is kept as small as possible while maintaining a stable training, i.e.  avoiding large training times or non-numerical outputs.
\end{itemize}

Once $\sigma$ is set, we study how the median value of the NPLM test distribution under the null hypothesis varies with the number of Nystrom centers ($M$) and the regularization parameter ($\lambda$). 
We note that the median value of the empirical distribution of the test has a saturation region along $M$, i.e. from a certain value of $M$ the median is almost constant. We use this information to select $M$. Both values of M and $\lambda$ have to deal with the timing performance for a single toy which has to be reasonable. 
Thus we select the value of $\lambda$ as small as possible to preserve the kernel method complexity while keeping the execution time reasonable.
Lastly, it is fundamental to consider the compatibility of the resulting empirical distribution $p(t|H_0)$ with a $\chi^2$ distribution. A p-value for the Kolmorogov-Smirnov test is estimated by comparing the t distribution with the fitted $\chi^2$, in order to ensure the trend of the distribution $p(t|H_0)$. This aspect is important when the final p-value is estimated.